\documentclass[10pt,twocolumn,letterpaper]{article}

\usepackage{cvpr}
\cvprfinaltrue
\usepackage{times}
\usepackage{epsfig}
\usepackage{graphicx}
\usepackage{amsmath}
\usepackage{amssymb}
\usepackage{adjustbox}
\usepackage{threeparttable}
\usepackage{xcolor}
\usepackage{authblk}

\usepackage[pagebackref=true,breaklinks=true,letterpaper=true,colorlinks,bookmarks=false]{hyperref}

\begin{document}

\title{Extremely Dense Point Correspondences using a Learned Feature Descriptor} %
\author[1]{Xingtong Liu}
\author[1]{Yiping Zheng}
\author[1]{Benjamin Killeen}
\author[2]{Masaru Ishii}
\author[1]{Gregory D. Hager}
\author[1]{Russell H. Taylor}
\author[1]{Mathias Unberath}
\affil[1]{Johns Hopkins University}
\affil[2]{Johns Hopkins Medical Institutions}
\affil[ ]{\tt\small \{xingtongliu, unberath\}@jhu.edu}

\maketitle

\begin{abstract}
High-quality 3D reconstructions from endoscopy video play an important role in many clinical applications, including surgical navigation where they enable direct video-CT registration. While many methods exist for general multi-view 3D reconstruction, these methods often fail to deliver satisfactory performance on endoscopic video. Part of the reason is that local descriptors that establish pair-wise point correspondences, and thus drive reconstruction, struggle when confronted with the texture-scarce surface of anatomy. Learning-based dense descriptors usually have larger receptive fields enabling the encoding of global information, which can be used to disambiguate matches. In this work, we present an effective self-supervised training scheme and novel loss design for dense descriptor learning. In direct comparison to recent local and dense descriptors on an in-house sinus endoscopy dataset, we demonstrate that our proposed dense descriptor can generalize to unseen patients and scopes, thereby largely improving the performance of Structure from Motion (SfM) in terms of model density and completeness. We also evaluate our method on a public dense optical flow dataset and a small-scale SfM public dataset to further demonstrate the effectiveness and generality of our method. The source code is available at \url{https://github.com/lppllppl920/DenseDescriptorLearning-Pytorch}.
\end{abstract}

\section{Introduction}
\textbf{Background.}\quad In computer vision, correspondence estimation aims to find a match between 2D points in image space and corresponding 3D locations. Many potential applications rely on this fundamental task, such as Structure from Motion (SfM), Simultaneous Localization and Mapping (SLAM), image retrieval, and image-based localization. In particular, SfM and SLAM have been shown to be effective for endoscopy-based surgical navigation~\cite{Mahmoud2016ORBSLAMBasedET}, video-CT registration~\cite{leonard2018evaluation}, and lesion localization~\cite{Widya20193DRO}. These successes rely on the fact that SfM and SLAM estimate a sparse 3D structure of the observed scene as well as the camera's trajectory from unlabeled video, simultaneously. 

The advantages of SLAM and SfM are complementary. In applications that require real-time estimation, \emph{e.g.}~surgical navigation, SLAM provides a computationally efficient framework for correspondence estimation. Robust camera tracking requires a dense 3D reconstruction estimated from previous frames, but computational constraints usually limit SLAM to local optimization. This often leads to drifting errors, especially when the trajectory loop is not evident. On the other hand, SfM prioritizes high density and accuracy for the sparse 3D structure. This is due to the time-consuming global optimization used in the bundle adjustment, which limits SfM to applications where offline estimation is acceptable.

\begin{figure*}[]
	\centering
	\includegraphics[width=172mm]{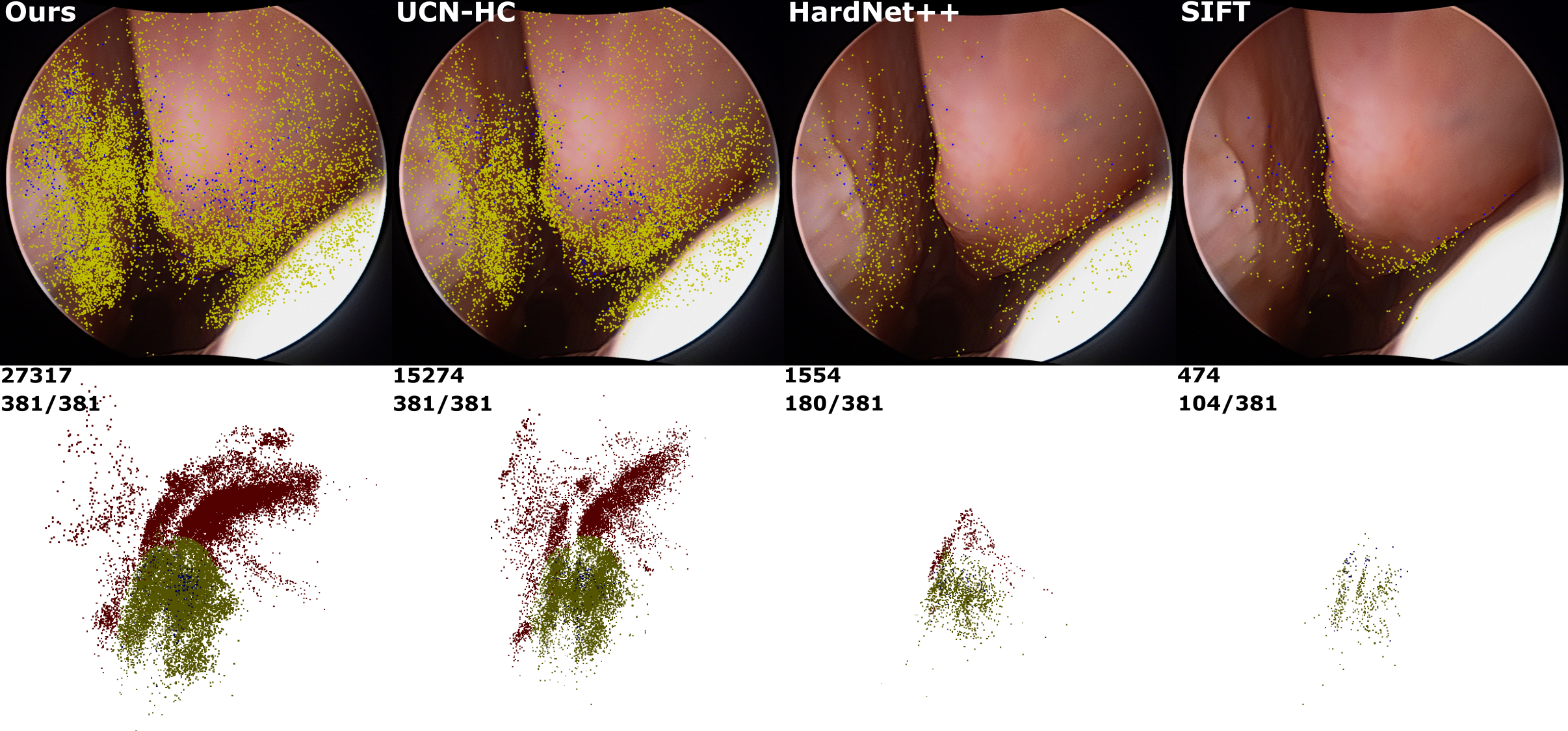}
	\caption{
	    \textbf{Qualitative comparison of SfM performance in endoscopy.}\quad The figure shows the performance of different descriptors on the task of SfM on the same sinus endoscopy video sequence. The comparison descriptors are ours, UCN~\cite{choy2016universal} trained with recently proposed Hardest Contrastive Loss~\cite{choy2019fully} on endoscopy data, HardNet++~\cite{mishchuk2017working} fined-tuned with endoscopy data, and SIFT~\cite{Lowe2004SIFT}. The first row shows the same video frame and the reprojection of the corresponding sparse 3D reconstruction from SfM. The second row displays the sparse reconstructions and relevant statistics; the number in the first row of each image is the number of points in the reconstruction; the two numbers in the second row of each image are the number of registered views and the total number of views in the sequence. The red points are those not visible in the showed frame. The yellow points are in the field of view of the displayed frame but reconstructed by other frames. The triangulation of blue points in the figure involves the displayed frame.	}
	\label{fig:point_cloud_overlay_comparison}

\end{figure*}

In video-CT registration, a markerless approach relies on correspondence estimation to provide a sparse reconstruction and the camera trajectory from the video. The reconstruction is then registered to the CT surface model with a registration algorithm~\cite{SINHA2019148}. This requires SfM since it relies on a dense and accurate 3D reconstruction. The accuracy of the estimated camera trajectory is also crucial so that the camera pose of each video frame aligns with the CT surface model. However, when estimating camera trajectory from endoscopic video, typical SfM or SLAM pipeline fails to produce a high-quality reconstruction or accurate camera trajectory. Recent work aims to mitigate this shortcoming through procedural changes in the video capture, which we discuss below.  In this work, we focus on developing a more effective feature descriptor, which is used in the feature extraction and matching module of the pipeline, to substantially increase the density of extracted correspondences (cf.~Fig.~\ref{fig:point_cloud_overlay_comparison}).

\textbf{Related Work.}\quad A local descriptor consists of a feature vector computed from an image patch, whose size and orientation are usually determined by a keypoint detector, such as Harris~\cite{Harris88acombined}, FAST~\cite{Rosten2006FAST}, and DoG~\cite{Lowe2004SIFT}. The hand-crafted local descriptor SIFT~\cite{Lowe2004SIFT} has been arguably the most popular feature descriptor for correspondence estimation and related tasks. In recent years, advanced variants of SIFT have been proposed, such as RootSIFT~\cite{Arandjelovic2012RootSIFT}, RootSIFT-PCA~\cite{Bursuc2015RootSIFTPCA}, and DSP-SIFT~\cite{Dong2014DomainsizePI}. Some of these outperform the SIFT descriptor in tasks such as fundamental matrix estimation~\cite{bian2019evaluation}, pair-wise feature matching, and multi-view reconstruction~\cite{schonberger2017comparative}. Additionally, learning-based local descriptors have grown in popularity with the advent of deep learning, with recent examples being L2-Net~\cite{tian2017l2}, GeoDesc~\cite{luo2018geodesc}, and HardNet~\cite{mishchuk2017working}. Though learning-based methods have outperformed hand-crafted ones in many areas of computer vision, advanced variants of SIFT continue to perform on par with or better than their learning-based local descriptors~\cite{bian2019evaluation,schonberger2017comparative}.

Several dense descriptors have been proposed, such as DAISY~\cite{tola2009daisy}, UCN~\cite{choy2016universal}, and POINT\textsuperscript{2}~\cite{liao2019multiview}. Compared with local descriptors, which follow a \textit{detect-and-describe} approach~\cite{dusmanu2019d2}, dense descriptors extract image information without using a keypoint detector to find specific locations for feature extraction. As a result, dense descriptors have higher computation efficiency than local descriptors in applications that require dense matching. They also avoid the possibility of repeated keypoint detection~\cite{dusmanu2019d2}. On the other hand, learning-based dense descriptors typically show better performance compared with hand-crafted ones. This is because Convolutional Neural Networks (CNN) can encode and fuse high-level context and low-level texture information more effectively than manual rules given enough training data. Our method belongs to the category of learning-based dense descriptors. There are also works that jointly learn a dense descriptor and a keypoint detector, such as SuperPoint~\cite{detone2018superpoint} and D2-Net~\cite{dusmanu2019d2}, or learn a keypoint detector that improves the performance of a local descriptor, such as GLAMpoints~\cite{truong2019glampoints}.

In the field of endoscopy, researchers have applied SfM and SLAM to video from various anatomy, including sinus~\cite{leonard2018evaluation}, stomach~\cite{Widya20193DRO}, abdomen~\cite{grasa2013visual,Mahmoud2016ORBSLAMBasedET}, and oral cavity~\cite{qiu2018endoscope}. Popular SfM pipelines such as COLMAP~\cite{schoenberger2016sfm} and SLAM systems such as ORB-SLAM~\cite{mur2015orb} usually do not achieve satisfactory results in endoscopy without further improvement. Several challenges stand in the way of successful correspondence estimation in endoscopic video. First, tissue deformation, as in video from a colonoscopy, violates the static scene assumption in these pipelines. To mitigate this issue, researchers have proposed SLAM-based methods that tolerate scene deformation~\cite{lamarca2019defslam,song2018mis}. Second, the textures in endoscopy are often smooth and repetitive, which makes the sparse matching with local descriptors error-prone.  Widya~\etal~\cite{Widya20193DRO} proposed spreading IC dye in the stomach to manually add texture to the surface, increasing the matching performance of local descriptors. This leads to denser and more complete reconstructions. Qiu~\etal~\cite{qiu2018endoscope} use a laser projector to project patterns on the surface of the oral cavity to add more textures to improve the performance of a SLAM system. However, introducing additional procedures as above is usually not desired by surgeons because it will interrupt the original workflow. Therefore, instead of adding textures, we develop a dense descriptor that works well on the texture-scarce surface to replace the original local descriptors in these systems.

\textbf{Contributions.}\quad First, to the best of our knowledge, this is the first work that applies learning-based dense descriptors to the task of multi-view reconstruction in endoscopy. Second, we present an effective self-supervised training scheme which includes a novel loss called Relative Response Loss that can train a high-precision dense descriptor with the learning style of keypoint localization. The proposed training scheme outperforms the popular hard negative mining strategy used in various learning-based descriptors~\cite{choy2016universal,choy2019fully,mishchuk2017working}. For evaluation, we have conducted extensive comparative studies on the task of pair-wise feature matching and SfM on a sinus endoscopy dataset, pair-wise feature matching on the KITTI Flow 2015 dataset~\cite{Menze2015object}, and SfM on a small-scale natural scene dataset~\cite{strecha2008benchmarking}.

\section{Methods}
In this section, we describe our self-supervised training scheme for dense descriptor learning, which includes overall network architecture, training scheme, custom layers, loss design, and a dense feature matching method.

\begin{figure*}[]
	\centering
	\includegraphics[width=160mm]{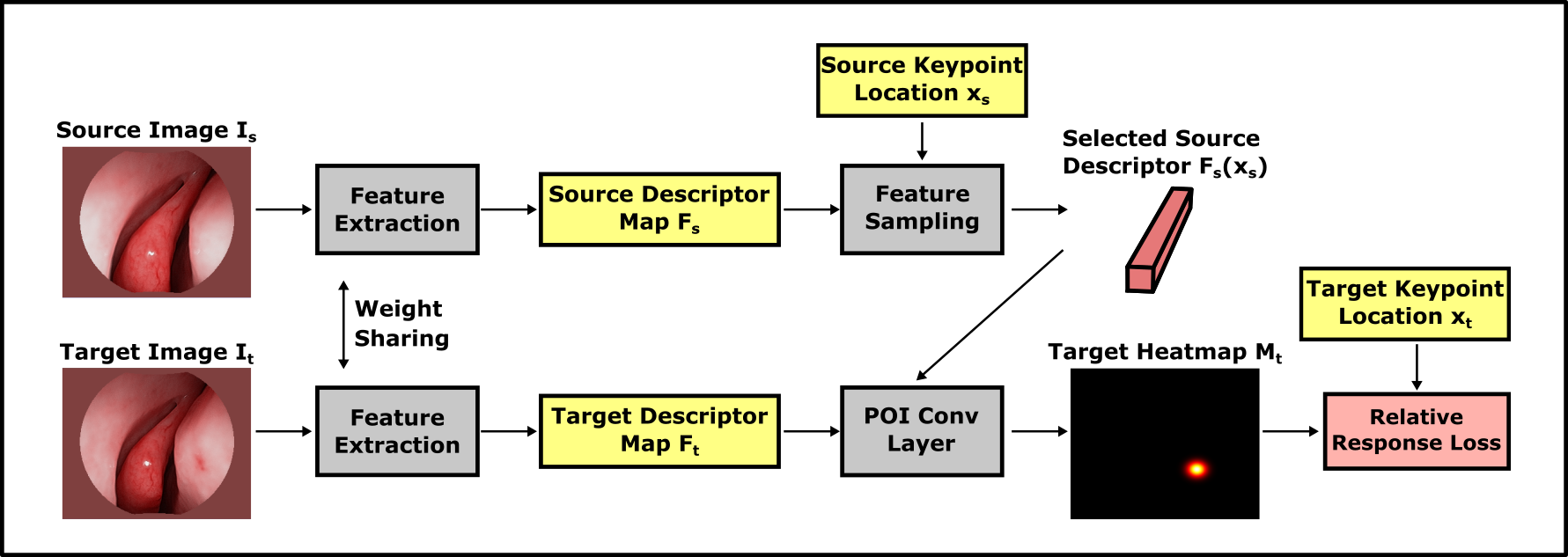}
	\caption{
	    \textbf{Overall network architecture.}\quad The training data consists of a pair of source and target images and groundtruth source-target 2D point correspondences. The source and target images are randomly selected from the frames which share observations of the same 3D points. For each pair of images, a certain number of point correspondences are randomly selected from the available ones in each training iteration. For the simplicity of illustration, only one target-source point pair and the corresponding target heatmap are shown in the figure. All concepts in the figure are defined in the \textit{Methods} section.
	}
	\label{fig:overall_network_architecture}
\end{figure*}

\textbf{Overall Network Architecture.}\quad As shown in Fig.~\ref{fig:overall_network_architecture}, the training network is a two-branch Siamese network. The input is a pair of color images, which are used as source and target. The training goal is, given a keypoint location in the source image, finding the correct corresponding keypoint location in the target image. A SfM method~\cite{leonard2018evaluation} with SIFT is applied to video sequences to estimate the sparse 3D reconstructions and camera poses. The groundtruth point correspondences are then generated by projecting the sparse 3D reconstructions onto the image planes using the estimated camera poses. The dense feature extraction module is a fully convolutional DenseNet~\cite{jegou2017one} which takes in a color image and output a dense descriptor map that has the same resolution as the input image and the length of the feature descriptor as the channel dimension. The descriptor map is L2-normalized along the channel dimension to increase the generalizability~\cite{wang2017normface}. For each source keypoint location, the corresponding descriptor is sampled from the source descriptor map. Using the descriptor of the source keypoint as a $1\times 1$ convolution kernel, a 2D convolution is performed on the target descriptor map in the Point-of-interest (POI) Conv Layer~\cite{liao2019multiview}. The computed heatmap represents the similarity between the source keypoint location and every location on the target image. The network is trained with proposed Relative Response Loss (RR) to force the heatmap to present high response only at the groundtruth target location. The idea of converting the problem of descriptor learning to keypoint localization is proposed by Liao~\etal~\cite{liao2019multiview}, which was originally used to solve the problem of X-ray-CT 2D-3D registration.

\textbf{Point-of-Interest (POI) Conv Layer.}\quad This layer is used to convert the problem of descriptor learning to keypoint localization~\cite{liao2019multiview}. For a pair of source and target input images, a pair of dense descriptor maps, $\mathbf{F}_\mathrm{s}$ and $\mathbf{F}_\mathrm{t}$, are generated from the feature extraction module. The size of an input image and a descriptor map are $3\times H\times W$ and $C\times H\times W$, respectively. For a descriptor at the source keypoint location $\mathbf{x}_{\mathrm{s}}$, the corresponding feature descriptor, $\mathbf{F}_\mathrm{s}\left(\mathbf{x}_{\mathrm{s}}\right)$, is extracted with the nearest neighbor sampling, which could be changed to other sampling methods if needed. The size of the descriptor is $C\times 1\times 1$. By treating the sampled feature descriptor as a $1\times1$ convolution kernel, the 2D convolution operation is performed on $\mathbf{F}_\mathrm{t}$ to generate a target heatmap, $\mathbf{M}_\mathrm{t}$, storing the similarity between the source descriptor and every target descriptor in $\mathbf{F}_\mathrm{t}$.

\begin{figure*}[]
	\centering
	\includegraphics[width=172mm]{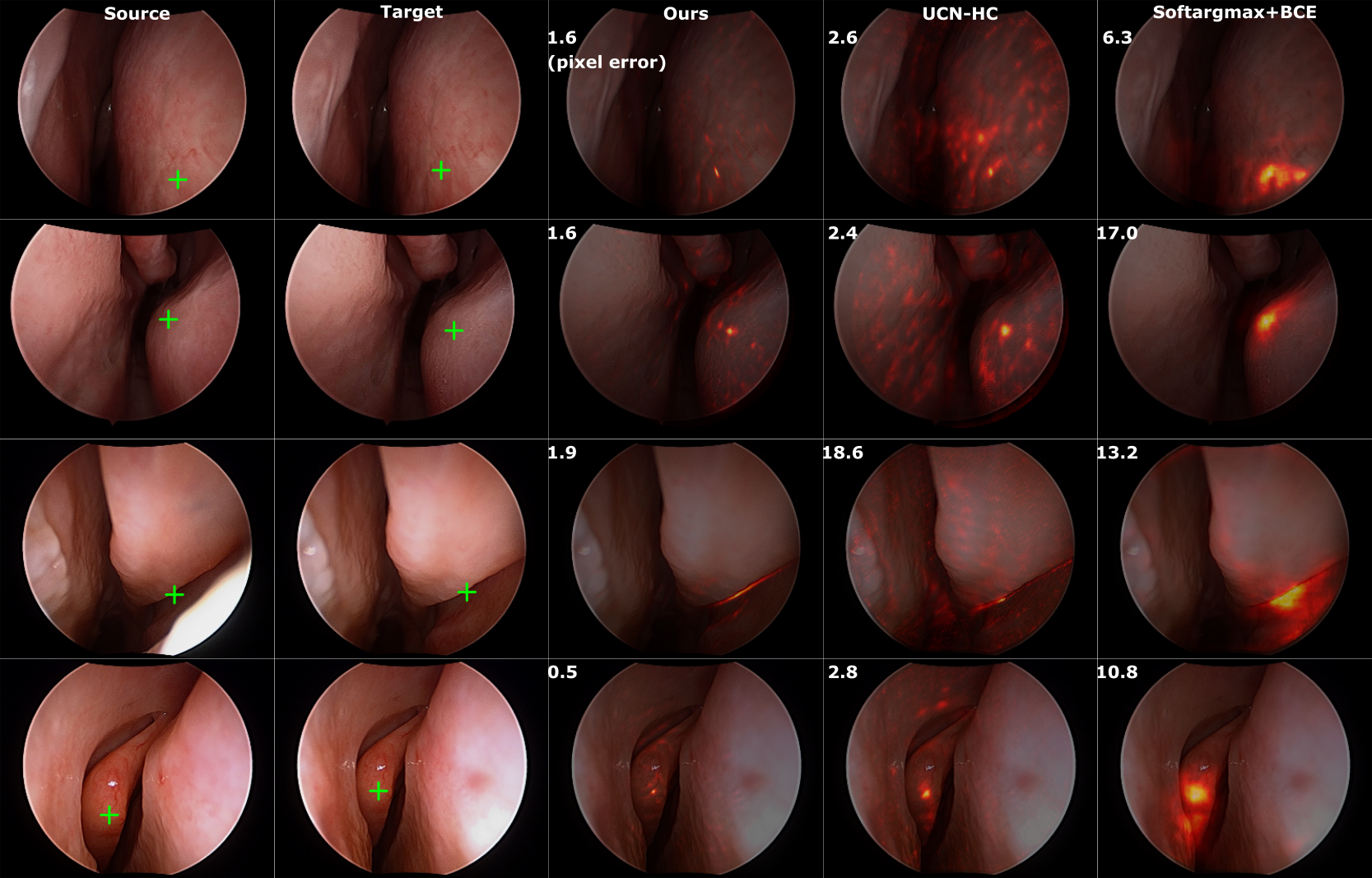}
	\caption{
	    \textbf{Qualitative comparison of feature matching performance in endoscopy.}\quad The figure qualitatively shows the performance of three dense descriptors trained with different loss designs on the task of pair-wise feature matching. The first two rows are training images and the rest are testing ones. The first and second columns show the source-target image pairs, where the green crossmarks indicate the groundtruth source-target point correspondences. For each dense descriptor, a target heatmap, as shown in the last three columns, is generated from the POI Conv Layer. To visualize the contrast better, the displayed heatmap is normalized with spatial softmax operation and then with the maximum value of the processed heatmap. The numbers shown in the last three columns are the pixel errors between the estimated target keypoint locations and the groundtruth ones. The fourth column shows the results of UCN~\cite{choy2016universal} trained with recent Hardest Contrastive Loss on the endoscopy dataset. The model in the fifth column is trained with the same method as ours except that the training loss is Softargmax~\cite{honari2018improving} and BCE instead of the proposed Relative Response Loss. The results show that our method produces fewer high responses, which leads to better matching accuracy.
	}
	\label{fig:matching_heatmap_comparison}
\end{figure*}

\textbf{Relative Response Loss (RR).}\quad The loss is proposed with the intuition that a target heatmap should present a high response at the groundtruth target keypoint location and the responses at other locations should be suppressed as much as possible. Besides, we do not want to assume any prior knowledge on the response distribution of the heatmap to preserve the potential of multimodal distribution to respect the matching ambiguity of challenging cases. To this end, we propose to maximize the ratio between the response at the groundtruth location and the summation of all responses of the heatmap. Mathematically, it is defined as,
\begin{equation}
\mathcal{L}_\mathrm{rr} = -\log\left( \frac{\mathrm{e}^{\sigma\mathbf{M}_\mathrm{t}\left(\mathbf{x}_{\mathrm{t}}\right)}}{\sum_{\mathbf{x}}\mathrm{e}^{\sigma\mathbf{M}_\mathrm{t}\left(\mathbf{x}\right)}}\right) \quad \text{, where}
\end{equation}
a scale factor $\sigma$ is applied to the heatmap $\mathbf{M}_\mathrm{t}$ to enlarge the value range which was $\left[-1, 1\right]$. A spatial softmax is then calculated at the groundtruth location $\mathbf{x}_{\mathrm{t}}$ of the scaled heatmap, where the denominator is the summation of all elements of the scaled heatmap. The logarithm operation is used to speed up the convergence. We observe that, by only penalizing the value at the groundtruth location after spatial softmax operation, the network learns to reduce the response at the other locations and increase the response at the groundtruth location effectively. We compare the feature matching and SfM performance of dense descriptors trained with different common loss designs that are originally for the task of keypoint localization in the \textit{Experiments} section. A qualitative comparison of target heatmaps generated by different dense descriptors is shown in Fig.~\ref{fig:matching_heatmap_comparison}.

\textbf{Dense Feature Matching.} For each source keypoint location in the source image, a corresponding target heatmap is generated with the method above. The location with the largest response value in the heatmap is selected as the estimated target keypoint location. The descriptor at the estimated target keypoint location then performs the same operation on the source descriptor map to estimate the source keypoint location. Because of the characteristics of dense matching, the traditional mutual nearest neighbor criterion used in the pair-wise feature matching of a local descriptor is too strict. We relax the criterion by accepting the match as long as the estimated source keypoint location is within the vicinity of the original source keypoint location, which we call cycle consistency criterion. The computation of dense matching can be parallelized on modern GPU by treating all sampled source descriptors as a kernel with the size of $N\times L\times 1\times 1$; $N$ is the number of query source keypoint locations used as the output channel dimension; $L$ is the length of the feature descriptor used as the input channel dimension of a standard 2D convolution operation.

\section{Experiments}
We evaluate our proposed method on three datasets. Sinus endoscopy dataset is used to evaluate the performance of local and dense descriptors on the task of pair-wise feature matching and SfM in endoscopy. KITTI Flow 2015 dataset~\cite{Menze2015object} is used to evaluate the performance of dense descriptors on the task of pair-wise feature matching in natural scenes. A small-scale dataset with a collection of building photos~\cite{strecha2008benchmarking} is used to evaluate the performance of local and dense descriptors on the task of SfM in natural scenes. All experiments are conducted on a workstation with $4$ NVIDIA Tesla M60 GPUs, each with $8$ GB memory, and the method is implemented using PyTorch~\cite{paszke2017automatic}. 

\begin{table}
\resizebox{\linewidth}{!}{%
\centering
\begin{tabular}{l|ccccccc}
\hline
 & \multicolumn{1}{l}{UCN-C} & \multicolumn{1}{l}{UCN-HC} & \multicolumn{1}{l}{Softarg.} & \multicolumn{1}{l}{Softarg.+BCE} & \multicolumn{1}{l}{Softmax+BCE} & \multicolumn{1}{l}{RR+Softarg.} & \multicolumn{1}{l}{RR} \\ \hline
PCK@5px & 25.5 & 58.8 & 36.5 & 44.6 & 35.4 & 57.9 & \textcolor{red}{63.0}\\
PCK@10px & 35.0 & 67.2 & 54.6 & 63.1 & 51.1 & 68.6 &  \textcolor{red}{71.9}\\
PCK@20px & 47.0 & 74.0 & 73.6 & 77.4 & 66.0 & 78.6 &  \textcolor{red}{80.0}\\ \hline
\end{tabular}%
}
\caption{\textbf{Evaluation of feature matching performance in endoscopy.}\quad This table shows the average percentage of correct keypoints (PCK) with threshold 5px, 10px, and 20px over all 9 sequences from the 3 testing patients. The PCK is calculated on all image pairs whose interval is within 20 frames. For each pair, PCK is computed by comparing the dense matching results with the groundtruth point correspondences from SfM results. The feature matching results in each column are generated by the descriptor whose name is on the first row. From left to right, the evaluated descriptors are UCN trained with Contrastive Loss (UCN-C)~\cite{choy2016universal}, UCN trained with Hardest Contrastive Loss (UCN-HC)~\cite{choy2019fully}, replacing the proposed Relative Response Loss (RR) with Softargmax~\cite{honari2018improving}, replacing RR with Softargmax and Binary Cross Entropy (BCE), replacing RR with spatial softmax and BCE~\cite{he2017mask}, RR and Softargmax, and the proposed training scheme with RR. The model trained with the proposed RR achieves the best average matching accuracy.}
\label{tab:feature_matching_evaluation_endoscopy}
\end{table}

\begin{table*}
\resizebox{\textwidth}{!}{%
\centering
\begin{tabular}{l|rrr|rrr|rrr|rrr|rrr|rrr|rrr|rrr|rrr}
\hline
\multicolumn{1}{l|}{} & \multicolumn{3}{l|}{Seq. 1-1 (381)}                                               & \multicolumn{3}{l|}{Seq. 1-2 (314)}      & \multicolumn{3}{l|}{Seq. 1-3 (370)} & \multicolumn{3}{l|}{Seq. 2-1 (455)} & \multicolumn{3}{l|}{Seq. 2-2 (630)} & \multicolumn{3}{l|}{Seq. 2-3 (251)} & \multicolumn{3}{l|}{Seq. 3-1 (90)} & \multicolumn{3}{l|}{Seq. 3-2 (1309)} & \multicolumn{3}{l}{Seq. 3-3 (336)} \\ \hline
SIFT & 104 & 474 & 5.62 & 219 & 1317 & 5.58 & 113 & 938 & \textcolor{red}{5.16} & 119 & 751 & 5.81 & 295 & 10384 & \textcolor{red}{6.43} & 122 & 1896 & \textcolor{red}{5.38} & 48 & 435  &  5.09  & 55 & 953 & \textcolor{red}{5.51} & 169 & 2169 & 5.57 \\
DSP-SIFT & 149 & 783 & 5.09 & 235 & 1918 & 5.06 & 132 & 1228 & 4.78 & 404 & 6557 & 5.32 & 296 & 7322 & 5.64 & 167 & 3450 & 5.00 & 42 & 293 & 4.81 & 150 & 745 & 5.17 & 180 & 1180 & 5.18 \\
RootSIFT-PCA & 104 & 384 & \textcolor{red}{5.89} & 219 & 1004 & \textcolor{red}{5.67} & 115 & 661 & 5.11 & 227 & 821 & \textcolor{red}{5.82} & 295 & 10147 & 6.43 & 128 & 2025 & 5.46 & 50 & 255 & \textcolor{red}{5.18} & 217 & 3188 & 5.35 & 176 & 2450 & \textcolor{red}{5.62} \\
HardNet++ & 180 & 1554 & 4.63 & 233 & 2162 & 4.81 & 244 & 3003 & 4.65 & 424 & 4755 & 4.65 & 534 & 9828 & 4.85 & 225 & 5727 & 4.56 & 79 & 610 & 4.66 & 416 & 4658 &   4.62   &   228    &   3196    &   4.66\\
UCN-C   &  349 &    13402 &         4.26              &        311              & 13198 & 4.50 &   248    &   8336    &    4.43   &   405    &   11935    &   4.13    &   293    &   8258    &   4.46    &   196    &   9273    &   3.98    &   77    &   2445    &    4.10   &    503   &   16166    &   4.29    &    206   &   3736    &  4.17    \\
UCN-HC                      &         \textcolor{red}{381}             &         15274             &         4.84              &         \textcolor{red}{314}             & 13519 & 4.84 &   352    &   16900    &   4.89    &   \textcolor{red}{455}    &   33299    &   4.67    &  \textcolor{red}{630}     &   45375    &    4.81   &   \textcolor{red}{251}    &   \textcolor{red}{26322}    &   4.37    &   86    &  2988     &    4.39   &   484    &  13394     &    4.39   &     283  &   11555    &   4.39   \\
Softarg.                      &348 &	5966 & 4.74 & 312 & 7774 & 4.74	 & 252	& 7426 & 4.63 & 293	& 4861 & 4.50 & 547 & 12590& 4.24 & 205	& 2847 & 4.22  & 59 &	534 & 4.17 	& 451 &	7247& 4.76 	& 302 &	6039 & 4.26 \\
Softarg.+BCE                      & 357	&	11502	&	4.47	& \textcolor{red}{314}	&	10373	&	4.57	& 244	&	10339	&	4.55	& 426	&	19848	&	4.34	& 560	&	22482	&	4.19	& 125	&	1150	&	4.04	& 46	&	774	&	4.04	& 500	&	12187	&	4.51	& 303	&	6268	&	4.06  \\
Softmax+BCE                      & 165	&	2246	&	4.26	& 306	&	8885	&	4.26	& 228	&	8628	&	4.19	& 378	&	8559	&	4.10	& 296	&	12081	&	4.19	& 77	&	1124	&	3.96	& 34	&	353	&	4.02	& 261	&	5024	&	4.19	& 181	&	2973	&	4.07 \\
RR+Softarg.                     & \textcolor{red}{381}	&	19921	&	4.99	& \textcolor{red}{314}	&	20375	&	4.98	& 256	&	20550	&	4.94	& \textcolor{red}{455}	&	\textcolor{red}{44388}	&	4.75	& \textcolor{red}{630}	& 39752	&	4.64	& 244	&	10055	&	4.35	& 87	&	5071	&	4.33	& \textcolor{red}{507}	&	20906	&	4.61	& 312	&	12856	&	4.36 \\
RR                      &  \textcolor{red}{381}	&	\textcolor{red}{27317}	&	5.07	&\textcolor{red}{314}	&	\textcolor{red}{22898}	&	5.23	&\textcolor{red}{367}	&	\textcolor{red}{29734}	&	5.06	&\textcolor{red}{455}	&	41380	&	4.78	&\textcolor{red}{630}	&	\textcolor{red}{45654}	&	4.80	&\textcolor{red}{251}	&	19645	&	4.43	&\textcolor{red}{89}	&	\textcolor{red}{6763}	&	4.62	&\textcolor{red}{507}	&	\textcolor{red}{35645}	&	4.68	&\textcolor{red}{313}	&	\textcolor{red}{21703}	&	4.53    \\ \hline
\end{tabular}
}
\caption{\textbf{Evaluation of SfM performance in endoscopy.}\quad We compare the SfM results of 9 sequences from the 3 testing patients. The SfM results are generated by the descriptors whose names are on the first column. We compare the SfM performance of local and dense descriptors. Starting from the first descriptor, these are SIFT~\cite{Lowe2004SIFT}, DSP-SIFT~\cite{Dong2014DomainsizePI}, RootSIFT-PCA~\cite{Bursuc2015RootSIFTPCA}, HardNet++~\cite{mishchuk2017working} fine-tuned with the endoscopy dataset, UCN trained with Contrastive Loss (UCN-C)~\cite{choy2016universal}, UCN trained with Hardest Contrastive Loss (UCN-HC)~\cite{choy2019fully}, replacing the proposed RR with Softargmax~\cite{honari2018improving}, replacing RR with Softargmax and BCE, replacing RR with spatial softmax and BCE~\cite{he2017mask}, RR and Softargmax, and the proposed training scheme with RR. Each number in the first row represents the number of frames in each sequence. In the following rows, for each sequence and each method, three numbers from left to right are the number of registered views, the number of sparse points, and the average track length of sparse points. It shows that the proposed method (RR) obtains the most number of registered views in all sequences and the densest reconstructions for most of the sequences. SIFT or RootSIFT-PCA achieve highest average track length in all sequences.}
\label{tab:sfm_evaluation_endoscopy}
\end{table*}

\textbf{Evaluation on Sinus Endoscopy.}\quad The dataset consists of video data collected from 8 patients and 2 cadavers. The overall duration is around 30 minutes. For the ease of experiments, all images are downsampled to $256\times 320$ pixels during both training and testing. For our method, we use a light-weight version of Fully Convolutional DenseNet (FC-DenseNet) ~\cite{jegou2017one} with 32 layers and filter growth rate of 10. The length of the output descriptor is 256; the overall number of parameters is 0.53 million. The model is trained with Stochastic Gradient Descent with the cyclic learning rate~\cite{smith2017cyclical} ranging from 1.0e-4 to 1.0e-3. The scale factor $\sigma$ used in the Relative Response Loss is empirically set to 20.0. Data from 5 patients and 1 cadaver are used for training; the other cadaver is used for validation; the remaining 3 patients are for testing. Because our evaluation focuses on the loss design, for fairness, we use the same network architecture described above for all dense descriptors to extract features. All models are trained until the performance on the validation data stops improving. The evaluation results of pair-wise feature matching are shown in Table~\ref{tab:feature_matching_evaluation_endoscopy}. To measure the accuracy of feature matching, we use the percentage of correct keypoints (PCK) with three thresholds, which are 5, 10 and 20 pixels. The matching is determined as correct if the detected target keypoint location is within a specified number of pixels. The results show that our proposed training scheme for the dense descriptor outperforms competing methods for dense descriptor learning, which are Contrastive Loss in~\cite{choy2016universal} and Hardest Contrastive Loss in~\cite{choy2019fully}. Besides, since we convert the problem of descriptor learning to keypoint localization, we also evaluate the performance of several loss functions used in keypoint localization by training the proposed network with these instead of Relative Response Loss. For the proposed method, generating and matching a pair of dense descriptor maps under the current setting takes about $37\mathrm{ms}$. To evaluate the performance of local and dense descriptors on the task of SfM in endoscopy, we use a simple SfM pipeline~\cite{leonard2018evaluation} which takes in pair-wise feature matches, uses Hierarchical Multi-Affine~\cite{puerto2012hierarchical} for geometric verification, and global bundle adjustment~\cite{moulon2016openmvg} for optimization. Pair-wise feature matches are estimated in all image pairs whose interval is within 30 frames. For all local descriptors, DoG~\cite{Lowe2004SIFT} is used to extract keypoint locations in both source and target images for sparse feature matching with mutual nearest neighbor (MNN) as the matching criterion. For dense descriptors, DoG is used to extract keypoint locations in only source images and dense matching is performed on the target images for these detected candidate keypoint locations in the source images. The false matches are ruled out using the cycle consistency criterion described in the \textit{Dense Feature Matching} subsection. Because of the texture smoothness of endoscopy, we change the hyperparameters of DoG so that more candidate keypoint locations can be detected. The number of layers in each octave is 8; the contrast threshold is $5.0\mathrm{e}^{-5}$; the edge threshold is 100.0; the standard deviation of the Gaussian applied to the image at the first octave is 1.1. All hand-crafted descriptors use the parameter setting recommended by the original authors. The SfM results are shown in Table~\ref{tab:sfm_evaluation_endoscopy}. Note that we build an image patch dataset from SfM results in endoscopy using the same method as~\cite{luo2018geodesc} to fine-tune the HardNet++~\cite{mishchuk2017working} for a fair comparison, which does have better performance compared with the pre-trained model released by the authors. 

\begin{figure}
	\centering
	\includegraphics[width=\linewidth]{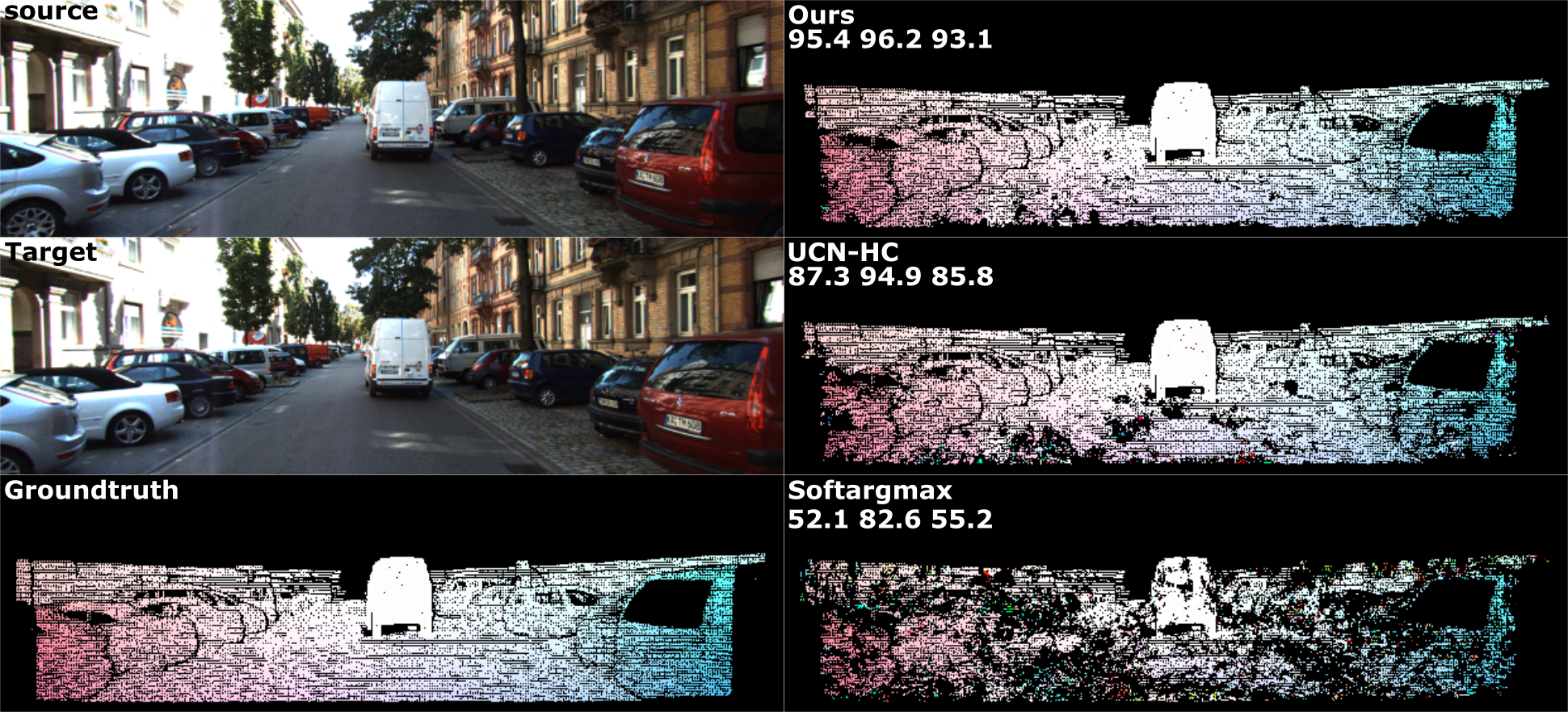}
	\caption{
	    \textbf{Qualitative comparison of feature matching performance in KITTI Flow 2015~\cite{Menze2015object}.}\quad The optical flow estimates from three dense descriptors for the same source-target image pair are shown in this figure. The descriptors are our method, UCN-HC~\cite{choy2019fully}, and the proposed method trained with Softargmax~\cite{honari2018improving} instead of RR.
	    The numbers shown in each image in the second column represent the putative match ratio, precision, and matching score of the optical flow estimates. We use the cycle consistency criterion with 6px threshold to rule out potential false matches. The images in the first column are source, target, and groundtruth dense optical flow map, where black values mean there are no valid measurements. The second column shows the dense optical flow estimates, where the black pixels include those with no groundtruth measurements or ruled out with cycle consistency criterion.  
	}
	\label{fig:kitti_flow_comparison}
\end{figure}

\textbf{Evaluation on KITTI Flow 2015~\cite{Menze2015object}.}\quad In this evaluation, we evaluate the performance of dense descriptors on the task of optical flow estimation. First, we estimate the $\text{Matching Score}=\frac{\text{\#Inliner Matches}}{\text{\#Features}}$. The $\#\text{Inlier Matches}$ is the number of matches where the distance between the estimated target keypoint location and the groundtruth target location is within 10 pixels. The $\#\text{Features}$ is equal to the number of pixels in an image. We also evaluate the $\text{Putative Match Ratio}=\frac{\text{\#Putative Matches}}{\text{\#Features}}$ and $\text{Precision}=\frac{\text{\#Inlier Matches}}{\text{\#Putative Matches}}$~\cite{schonberger2017comparative}.
The match is determined as putative if it passes the cycle consistency criterion. We follow the same training protocol as~\cite{choy2016universal}, where 1000 point correspondences are randomly selected for each image pair in the KITTI dataset and fixed during training. The models trained with the proposed Relative Response Loss, Softargmax loss~\cite{honari2018improving}, Contrastive Loss~\cite{choy2016universal}, and Hardest Contrastive Loss~\cite{choy2019fully}, respectively, are evaluated. To evaluate the performance of different loss designs, we train all models with the same network architecture for feature extraction. We use FC-DenseNet with 38 layers and filter growth rate of 16. The overall number of parameters is 1.68 million. Other parameter settings are the same as the evaluation in endoscopy. The images are downsampled by a factor of 2 during training. Two top-performing results of the comparison methods presented in~\cite{choy2016universal} are cited here. An example of dense optical flow estimates from different trained models is shown in Fig.~\ref{fig:kitti_flow_comparison}. The quantitative evaluation results are shown in Table~\ref{tab:kitti_flow_evaluation}.

\begin{table}

\centering
\begin{adjustbox}{max width=\linewidth}
\begin{tabular}{l|rrrrrrrrrr}
\hline
 & \multicolumn{1}{l}{DaisyFF} & \multicolumn{1}{l}{DM} & \multicolumn{1}{l}{UCN-C} & \multicolumn{1}{l}{UCN-HC} & \multicolumn{1}{l}{Softargmax} & \multicolumn{1}{l}{RR} \\ \hline
Putative Match Ratio (\%) & \textcolor{red}{100.0} & \textcolor{red}{100.0} & 73.6 & 80.6 & 46.0 & 88.0 \\
Precision (\%) & 79.6 & 85.6 & 73.2 & \textcolor{red}{90.9} & 70.7 & 89.8 \\
Matching Score (\%) & 79.6 & \textcolor{red}{85.6} & 61.9 & 76.9 & 41.5 & 81.7 \\ \hline
\end{tabular}%
\end{adjustbox}
\caption{\textbf{Evaluation of optical flow estimation on KITTI Flow 2015~\cite{Menze2015object}.}\quad DaisyFF~\cite{yang2014daisy} and DM~\cite{revaud2016deepmatching} use global optimization to estimate a dense optical flow map between a pair of images. The last four methods are introduced in Table~\ref{tab:sfm_evaluation_endoscopy}. It shows that, by removing unconfident matches using cycle consistency criterion with 6 pixels as the threshold, our method achieves marginally lower precision than UCN-HC, whereas UCN-HC has lower putative match ratio than ours. Our method obtains the highest matching score among the last four methods. Note that we assume the first two methods do not discard any matches, which is why the putative match ratios are shown as 100. }
\label{tab:kitti_flow_evaluation}
\end{table}

\textbf{Evaluation on Multi-view Stereo 2008~\cite{strecha2008benchmarking}.}\quad The dataset consists of several small-scale sequences where the same building is captured from different viewpoints within each sequence. We evaluate the performance of the proposed method on the task of SfM in natural scenes and compare it with the hand-crafted local descriptors. Our model is trained with the SfM results of \textit{gerrard-hall}, \textit{personal-hall}, and \textit{south-building}, which were released by the author of COLMAP~\cite{schoenberger2016sfm}. We use FC-DenseNet with 32 layers and filter growth rate as 16. The overall number of parameters is 1.26 million. Other parameter settings are the same as the evaluation in endoscopy. All images for training and testing are downsampled to $256\times 320$. All descriptors use the DoG keypoint detector with the same parameter setting as the evaluation in endoscopy. The evaluation results are shown in Table~\ref{tab:small_scale_sfm}. Most experiments are conducted on the same SfM pipeline~\cite{leonard2018evaluation} as in endoscopy. SIFT and DSP-SIFT are also evaluated with COLMAP.

\begin{table}
\centering
\begin{adjustbox}{max width=\linewidth}
\begin{tabular}{l|rrr|rrr|rrr|rrr}
\hline
 & \multicolumn{3}{l|}{Entry (10)} & \multicolumn{3}{l|}{Fountain (11)} & \multicolumn{3}{l|}{HerzJesu (25)} & \multicolumn{3}{l}{Castle (30)} \\ \hline
SIFT-COLMAP & \textcolor{red}{10} & 1557 & 3.86 & \textcolor{red}{11} & 1566 & 4.33 & \textcolor{red}{25} & 3637 & \textcolor{red}{5.91} & \textcolor{red}{30} & 3718 & 4.57 \\
DSP-SIFT-COLMAP & \textcolor{red}{10} & 1849 & \textcolor{red}{3.93} & \textcolor{red}{11} & 1769 & \textcolor{red}{4.42} & \textcolor{red}{25} & 3650 & 5.90 & \textcolor{red}{30} & 4203 & \textcolor{red}{4.78} \\
SIFT & \textcolor{red}{10} & 1444 & 3.54 & 10 & 2775 & 3.63 & 24 & 6706 & 4.59 & \textcolor{red}{30} & \textcolor{red}{4589} & 3.76 \\
DSP-SIFT & \textcolor{red}{10} & 3041 & 3.79 & \textcolor{red}{11} & 4244 & 3.85 & \textcolor{red}{25} & 7334 & 4.35 & 22 & 1804 & 3.96 \\
RootSIFT-PCA & 7 & 1109 & 3.42 & 10 & 2467 & 3.54 & \textcolor{red}{25} & 6584 & 4.47 & 21 & 3991 & 3.84 \\
RR & \textcolor{red}{10} & 2980 & 3.62 & 7 & 6293 & 3.72 & \textcolor{red}{25} & 12807 & 3.96 & 29 & 3684 & 3.67 \\
RR-SG & \textcolor{red}{10} & \textcolor{red}{5310} & 3.60 & 8 & \textcolor{red}{7676} & 3.90 & 24 & \textcolor{red}{15799} & 4.28 & 27 & 3431 & 3.88 \\ \hline
\end{tabular}
\end{adjustbox}
\caption{\textbf{Evaluation of SfM performance on Multi-view Stereo 2008~\cite{strecha2008benchmarking}.}\quad Though the scene variation between the training and testing dataset is large, our method (RR) still performs comparably against the hand-crafted local descriptors. However, compared with endoscopy, we do observe a larger amount of false matches in the pair-wise feature matching phase. This potentially means a dense descriptor needs a larger amount of training data or a limited receptive field to avoid overfitting when the scene variation is large. To test the hypothesis, we train another model, which is RR-SG, with 4 times smaller receptive field and gray image as input and similar number of parameters to RR. It shows that RR-SG produces denser reconstructions in three out of four sequences. This might mean, compared with RR-SG, RR overfits to the high-level context information to a larger degree. Compared with the SfM pipeline in~\cite{leonard2018evaluation}, COLMAP has more stable performance in terms of the completeness of the camera trajectory but usually smaller number of sparse points. This observation is similar to the results in~\cite{schoenberger2016sfm}, where they compared COLMAP with other SfM pipelines. It is probably due to the stricter inlier criteria of COLMAP.}
\label{tab:small_scale_sfm}
\end{table}

\section{Discussion}

\textbf{Intuition on the Performance Difference of Various Training Schemes for Dense Descriptor Learning.} \quad We attribute the performance difference between our method and UCN-HC to the different strategies of training data sampling. For UCN-HC, given a positive point pair, one hardest negative point is obtained in a minibatch for each of the points in the pair to calculate the negative loss. A diameter threshold is also set to avoid mining points that are too close to the positive point. A positive margin threshold and negative margin threshold is also set to avoid penalizing positive pairs that are close enough or negative pairs that are far enough. There are several potential problems with this setting. First, the strategy of hardest sample selection, which was also used similarly in the local descriptor training~\cite{mishchuk2017working}, could potentially lead to training instability, which was also mentioned by the original authors in their Github repository. Because for each iteration of training, only the hardest negative samples in a minibatch provide gradients to the network training with other samples being ignored, the gradient direction may not be helpful to these ignored samples. This could potentially lead to training oscillation where hardest samples jump among different samples but the network never converges to the optimal solution. The results of instability can be found in Fig.~\ref{fig:matching_heatmap_comparison}, where many high responses are scattered in the heatmap. Second, the manually specified diameter and margin thresholds could also lead to a suboptimal solution. Because samples that are within the diameter of a selected sample are not considered as negative ones, the network will never try to push nearby samples away from the selected one. Therefore, this limits the matching precision of the descriptor. This again can be observed in Fig.~\ref{fig:matching_heatmap_comparison}, where the high-response clusters around the groundtruth target locations appear to be wider than our proposed method. The margin thresholds in the loss design also remove the possibility of further pushing away negative samples from the positive ones and pulling positive pairs closer, which could be another reason for obtaining such heatmaps. As a comparison, in our method, for each sampled point in the source image, all points in the target image are observed in one training iteration. Only the groundtruth target point is considered a positive point and all other points are considered negative ones. This avoids the oscillation related to the descriptor distance between the selected source point and all points in the target image. The reason why this training scheme will not suffer from the problem of data imbalance is due to the proposed Relative Response Loss (RR). The goal of RR is to make the ratio between the response at the groundtruth target location and the summation of all responses in the target image as high as possible. By doing this, the network will try to suppress all responses except the one at the target groundtruth location. It does not assume any prior distribution of the response heatmap and conveys the goal of precise feature matching clearly, which we believe improves the expressivity of the network. 

We have also evaluated some common losses used in the task of keypoint localization, such as spatial softmax + BCE and Softargmax~\cite{honari2018improving}. Spatial softmax + BCE is used for heatmap regression so that the network produces a similar heatmap as the groundtruth one. However, because the groundtruth distribution is usually assumed to be Gaussian with a manually specified standard deviation, this limits the expressivity of the network in cases where Gaussian distribution is not optimal. This can be observed in the third row in Fig.~\ref{fig:matching_heatmap_comparison}, where the model trained with Softargmax + BCE tries to infer a Gaussian-like distribution around the groundtruth location. As a comparison, the learned descriptor in our proposed method naturally produces high response along the edge of the surface, which is where the most ambiguities come from. Besides, BCE also suffers from the data imbalance problem for the case where positive and negative samples are highly unbalanced, which is also observed in~\cite{lin2017focal}. Softargmax converts the task of keypoint localization to a position regression task where the network tries to produce a heatmap so that the centroid of the heatmap is close to the groundtruth target location. However, this suffers from the fact that any distribution where the centroid is equal to the target location will not be further penalized. Therefore, Softargmax makes the network easily trapped in suboptimal solutions of learning a discriminative descriptor, whereas there are no such training ambiguities in RR. Though this ambiguity can be reduced by combining Softargmax with BCE, the performance is still worse than RR, as observed in Table~\ref{tab:feature_matching_evaluation_endoscopy} and~\ref{tab:sfm_evaluation_endoscopy}, because of the unimodal distribution assumption.

\textbf{Local Descriptor vs. Learning-based Dense Descriptor.}\quad We observe that learning-based dense descriptors usually perform better than local descriptors in the experiments related to SfM in sinus endoscopy. We attribute this to two reasons. First, local descriptors usually need a keypoint detector, such as DoG~\cite{Lowe2004SIFT}, to detect candidate keypoints before sparse feature matching. The lack of repeatability in the keypoint detector makes many true matches unable to be found because either source or target locations for these matches are not detected as candidate keypoints in the keypoint detection phase. As observed in~\cite{dusmanu2019d2}, the unstable detection is because the detector usually uses low-level information, which is often significantly affected by changes such as viewpoint and illumination. Second, the smooth and repetitive textures in endoscopy make it challenging for the local descriptors that have a limited receptive field to find correct matches even if all points in the true matches are detected by the keypoint detector. On the other hand, learning-based dense descriptors do not rely on the keypoint detector to produce repeatable keypoint locations and have a larger receptive field. 

Compared with local descriptors, dense descriptors also have disadvantages. First, a dense descriptor is more memory-demanding. This is because, to parallelize the dense matching procedure with many keypoint locations, the descriptors need to be organized in the form described in the \textit{Dense Feature Matching} subsection. This requires memory to store a response target heatmap for each source keypoint location before the target location is estimated from the heatmap. Though a sparse matching can also be performed with a dense descriptor, the performance will degrade because of the reliance on a repeatable keypoint detector. Therefore, the practical usage of a dense descriptor on a low-cost embedded system is limited. Second, learning-based dense descriptors seem to be more overfitting-prone compared with learning-based local descriptors. This is because the dense descriptor network relies on both high-level and low-level image information to generate a descriptor map. Because high-level information, presumably, has more variation compared with low-level texture information that learning-based local descriptors only need, more training data is probably needed for a dense descriptor. The reason why dense descriptors seem to generalize well in endoscopy could be due to the lower anatomical variation compared with the variation in natural scenes. 

\section{Conclusion}
In this work, we propose an effective self-supervised training scheme with a novel loss design for the learning-based dense descriptor. To the best of our knowledge, this is the first work that applies a learning-based dense descriptor to endoscopy for multi-view reconstruction. We evaluate our method on both endoscopy and natural scene dataset on the task of pair-wise feature matching and SfM, where our proposed method outperforms other local and dense descriptors on a sinus endoscopy dataset and outperforms recent dense descriptors in a dense optical flow public dataset. The extensive comparison study helps to gain more insights on the difference between local and dense descriptors, and the effects of different loss designs on the overall performance of a dense descriptor. Because SfM is an offline method, it is not able to support real-time localization and mapping. We plan to extend this work to incorporate a learning-based dense descriptor into an existing SLAM system in the future to make it more accurate and robust for surgical navigation in endoscopy. We also plan to adopt a bootstrapping training method to train the dense descriptor because of the observation that a descriptor model trained with sparse SfM results helps the SfM estimate denser reconstructions from both testing and training sequences.

{\small
\bibliographystyle{ieee}
\bibliography{main}

\begin{thebibliography}{10}\itemsep=-1pt

\bibitem{Arandjelovic2012RootSIFT}
R.~Arandjelovic.
\newblock Three things everyone should know to improve object retrieval.
\newblock In {\em Proceedings of the 2012 IEEE Conference on Computer Vision
  and Pattern Recognition (CVPR)}, CVPR '12, pages 2911--2918, Washington, DC,
  USA, 2012. IEEE Computer Society.

\bibitem{bian2019evaluation}
J.-W. Bian, Y.-H. Wu, J.~Zhao, Y.~Liu, L.~Zhang, M.-M. Cheng, and I.~Reid.
\newblock An evaluation of feature matchers for fundamental matrix estimation.
\newblock In {\em British Machine Vision Conference (BMVC)}, 2019.

\bibitem{Bursuc2015RootSIFTPCA}
A.~Bursuc, G.~Tolias, and H.~J{\'e}gou.
\newblock Kernel local descriptors with implicit rotation matching.
\newblock In {\em Proceedings of the 5th ACM on International Conference on
  Multimedia Retrieval}, ICMR '15, pages 595--598, New York, NY, USA, 2015.
  ACM.

\bibitem{choy2019fully}
C.~Choy, J.~Park, and V.~Koltun.
\newblock Fully convolutional geometric features.
\newblock In {\em Proceedings of the IEEE International Conference on Computer
  Vision}, pages 8958--8966, 2019.

\bibitem{choy2016universal}
C.~B. Choy, J.~Gwak, S.~Savarese, and M.~Chandraker.
\newblock Universal correspondence network.
\newblock In {\em Advances in Neural Information Processing Systems}, pages
  2414--2422, 2016.

\bibitem{Dong2014DomainsizePI}
J.~Dong and S.~Soatto.
\newblock Domain-size pooling in local descriptors: Dsp-sift.
\newblock {\em 2015 IEEE Conference on Computer Vision and Pattern Recognition
  (CVPR)}, pages 5097--5106, 2014.

\bibitem{dusmanu2019d2}
M.~Dusmanu, I.~Rocco, T.~Pajdla, M.~Pollefeys, J.~Sivic, A.~Torii, and
  T.~Sattler.
\newblock D2-net: A trainable cnn for joint detection and description of local
  features.
\newblock In {\em Proceedings of the 2019 IEEE/CVF Conference on Computer
  Vision and Pattern Recognition}, 2019.

\bibitem{grasa2013visual}
O.~G. Grasa, E.~Bernal, S.~Casado, I.~Gil, and J.~Montiel.
\newblock Visual slam for handheld monocular endoscope.
\newblock {\em IEEE transactions on medical imaging}, 33(1):135--146, 2013.

\bibitem{Harris88acombined}
C.~Harris and M.~Stephens.
\newblock A combined corner and edge detector.
\newblock In {\em In Proc. of Fourth Alvey Vision Conference}, pages 147--151,
  1988.

\bibitem{he2017mask}
K.~He, G.~Gkioxari, P.~Doll{\'a}r, and R.~Girshick.
\newblock Mask r-cnn.
\newblock In {\em Proceedings of the IEEE international conference on computer
  vision}, pages 2961--2969, 2017.

\bibitem{honari2018improving}
S.~Honari, P.~Molchanov, S.~Tyree, P.~Vincent, C.~Pal, and J.~Kautz.
\newblock Improving landmark localization with semi-supervised learning.
\newblock In {\em Proceedings of the IEEE Conference on Computer Vision and
  Pattern Recognition}, pages 1546--1555, 2018.

\bibitem{jegou2017one}
S.~J{\'e}gou, M.~Drozdzal, D.~Vazquez, A.~Romero, and Y.~Bengio.
\newblock The one hundred layers tiramisu: Fully convolutional densenets for
  semantic segmentation.
\newblock In {\em Proceedings of the IEEE Conference on Computer Vision and
  Pattern Recognition Workshops}, pages 11--19, 2017.

\bibitem{lamarca2019defslam}
J.~Lamarca, S.~Parashar, A.~Bartoli, and J.~Montiel.
\newblock Defslam: Tracking and mapping of deforming scenes from monocular
  sequences.
\newblock {\em arXiv preprint arXiv:1908.08918}, 2019.

\bibitem{leonard2018evaluation}
S.~{Leonard}, A.~{Sinha}, A.~{Reiter}, M.~{Ishii}, G.~L. {Gallia}, R.~H.
  {Taylor}, et~al.
\newblock Evaluation and stability analysis of video-based navigation system
  for functional endoscopic sinus surgery on in vivo clinical data.
\newblock 37(10):2185--2195, Oct. 2018.

\bibitem{liao2019multiview}
H.~Liao, W.~Lin, J.~Zhang, J.~Zhang, J.~Luo, and S.~K. Zhou.
\newblock Multiview 2d/3d rigid registration via a point-of-interest network
  for tracking and triangulation.
\newblock In {\em {IEEE} Conference on Computer Vision and Pattern Recognition,
  {CVPR} 2019, Long Beach, CA, USA, June 16-20, 2019}, pages 12638--12647.
  Computer Vision Foundation / {IEEE}, 2019.

\bibitem{lin2017focal}
T.-Y. Lin, P.~Goyal, R.~Girshick, K.~He, and P.~Doll{\'a}r.
\newblock Focal loss for dense object detection.
\newblock In {\em Proceedings of the IEEE international conference on computer
  vision}, pages 2980--2988, 2017.

\bibitem{Lowe2004SIFT}
D.~G. Lowe.
\newblock Distinctive image features from scale-invariant keypoints.
\newblock {\em Int. J. Comput. Vision}, 60(2):91--110, Nov. 2004.

\bibitem{luo2018geodesc}
Z.~Luo, T.~Shen, L.~Zhou, S.~Zhu, R.~Zhang, Y.~Yao, T.~Fang, and L.~Quan.
\newblock Geodesc: Learning local descriptors by integrating geometry
  constraints.
\newblock In {\em Proceedings of the European Conference on Computer Vision
  (ECCV)}, pages 168--183, 2018.

\bibitem{Mahmoud2016ORBSLAMBasedET}
N.~Mahmoud, I.~Cirauqui, A.~Hostettler, C.~Doignon, L.~Soler, J.~Marescaux, and
  J.~M.~M. Montiel.
\newblock Orbslam-based endoscope tracking and 3d reconstruction.
\newblock In {\em CARE@MICCAI}, 2016.

\bibitem{Menze2015object}
M.~Menze and A.~Geiger.
\newblock Object scene flow for autonomous vehicles.
\newblock In {\em The IEEE Conference on Computer Vision and Pattern
  Recognition (CVPR)}, June 2015.

\bibitem{mishchuk2017working}
A.~Mishchuk, D.~Mishkin, F.~Radenovic, and J.~Matas.
\newblock Working hard to know your neighbor's margins: Local descriptor
  learning loss.
\newblock In {\em Advances in Neural Information Processing Systems}, pages
  4826--4837, 2017.

\bibitem{moulon2016openmvg}
P.~Moulon, P.~Monasse, R.~Perrot, and R.~Marlet.
\newblock Openmvg: Open multiple view geometry.
\newblock In {\em International Workshop on Reproducible Research in Pattern
  Recognition}, pages 60--74. Springer, 2016.

\bibitem{mur2015orb}
R.~Mur-Artal, J.~M.~M. Montiel, and J.~D. Tardos.
\newblock Orb-slam: a versatile and accurate monocular slam system.
\newblock {\em IEEE transactions on robotics}, 31(5):1147--1163, 2015.

\bibitem{paszke2017automatic}
A.~Paszke, S.~Gross, S.~Chintala, G.~Chanan, E.~Yang, Z.~DeVito, Z.~Lin,
  A.~Desmaison, L.~Antiga, and A.~Lerer.
\newblock Automatic differentiation in pytorch.
\newblock 2017.

\bibitem{puerto2012hierarchical}
G.~A. Puerto-Souza and G.~L. Mariottini.
\newblock Hierarchical multi-affine (hma) algorithm for fast and accurate
  feature matching in minimally-invasive surgical images.
\newblock In {\em 2012 IEEE/RSJ International Conference on Intelligent Robots
  and Systems}, pages 2007--2012. IEEE, 2012.

\bibitem{qiu2018endoscope}
L.~Qiu and H.~Ren.
\newblock Endoscope navigation and 3d reconstruction of oral cavity by visual
  slam with mitigated data scarcity.
\newblock In {\em Proceedings of the IEEE Conference on Computer Vision and
  Pattern Recognition Workshops}, pages 2197--2204, 2018.

\bibitem{revaud2016deepmatching}
J.~Revaud, P.~Weinzaepfel, Z.~Harchaoui, and C.~Schmid.
\newblock Deepmatching: Hierarchical deformable dense matching.
\newblock {\em International Journal of Computer Vision}, 120(3):300--323,
  2016.

\bibitem{Rosten2006FAST}
E.~Rosten and T.~Drummond.
\newblock Machine learning for high-speed corner detection.
\newblock In {\em Proceedings of the 9th European Conference on Computer Vision
  - Volume Part I}, ECCV'06, pages 430--443, Berlin, Heidelberg, 2006.
  Springer-Verlag.

\bibitem{schoenberger2016sfm}
J.~L. Sch\"{o}nberger and J.-M. Frahm.
\newblock Structure-from-motion revisited.
\newblock In {\em Conference on Computer Vision and Pattern Recognition
  (CVPR)}, 2016.

\bibitem{schonberger2017comparative}
J.~L. Schonberger, H.~Hardmeier, T.~Sattler, and M.~Pollefeys.
\newblock Comparative evaluation of hand-crafted and learned local features.
\newblock In {\em Proceedings of the IEEE Conference on Computer Vision and
  Pattern Recognition}, pages 1482--1491, 2017.

\bibitem{SINHA2019148}
A.~Sinha, S.~D. Billings, A.~Reiter, X.~Liu, M.~Ishii, G.~D. Hager, and R.~H.
  Taylor.
\newblock The deformable most-likely-point paradigm.
\newblock {\em Medical Image Analysis}, 55:148 -- 164, 2019.

\bibitem{smith2017cyclical}
L.~N. Smith.
\newblock Cyclical learning rates for training neural networks.
\newblock In {\em 2017 IEEE Winter Conference on Applications of Computer
  Vision (WACV)}, pages 464--472. IEEE, 2017.

\bibitem{song2018mis}
J.~Song, J.~Wang, L.~Zhao, S.~Huang, and G.~Dissanayake.
\newblock Mis-slam: Real-time large-scale dense deformable slam system in
  minimal invasive surgery based on heterogeneous computing.
\newblock {\em IEEE Robotics and Automation Letters}, 3(4):4068--4075, 2018.

\bibitem{strecha2008benchmarking}
C.~Strecha, W.~Von~Hansen, L.~Van~Gool, P.~Fua, and U.~Thoennessen.
\newblock On benchmarking camera calibration and multi-view stereo for high
  resolution imagery.
\newblock In {\em 2008 IEEE Conference on Computer Vision and Pattern
  Recognition}, pages 1--8. Ieee, 2008.

\bibitem{tian2017l2}
Y.~Tian, B.~Fan, and F.~Wu.
\newblock L2-net: Deep learning of discriminative patch descriptor in euclidean
  space.
\newblock In {\em Proceedings of the IEEE Conference on Computer Vision and
  Pattern Recognition}, pages 661--669, 2017.

\bibitem{tola2009daisy}
E.~Tola, V.~Lepetit, and P.~Fua.
\newblock Daisy: An efficient dense descriptor applied to wide-baseline stereo.
\newblock {\em IEEE transactions on pattern analysis and machine intelligence},
  32(5):815--830, 2009.

\bibitem{wang2017normface}
F.~Wang, X.~Xiang, J.~Cheng, and A.~L. Yuille.
\newblock Normface: l 2 hypersphere embedding for face verification.
\newblock In {\em Proceedings of the 25th ACM international conference on
  Multimedia}, pages 1041--1049. ACM, 2017.

\bibitem{Widya20193DRO}
A.~R. Widya, Y.~Monno, K.~Imahori, M.~Okutomi, S.~Suzuki, T.~Gotoda, and
  K.~Miki.
\newblock 3d reconstruction of whole stomach from endoscope video using
  structure-from-motion.
\newblock {\em 2019 41st Annual International Conference of the IEEE
  Engineering in Medicine and Biology Society (EMBC)}, pages 3900--3904, 2019.

\bibitem{yang2014daisy}
H.~Yang, W.-Y. Lin, and J.~Lu.
\newblock Daisy filter flow: A generalized discrete approach to dense
  correspondences.
\newblock In {\em Proceedings of the IEEE Conference on Computer Vision and
  Pattern Recognition}, pages 3406--3413, 2014.

\end{thebibliography}
}

\end{document}